\documentclass[10pt,twocolumn,letterpaper]{article}

\usepackage{iccv}
\usepackage{times}
\usepackage{epsfig}
\usepackage{graphicx}
\usepackage{amsmath}
\usepackage{amssymb}
\usepackage{microtype}
\usepackage{booktabs}
\usepackage{authblk}
\usepackage{algorithm}
\usepackage{algpseudocode}
\usepackage[accsupp]{axessibility} 

\usepackage[pagebackref=true,breaklinks=true,letterpaper=true,colorlinks,bookmarks=false]{hyperref}

\iccvfinalcopy 

\def\iccvPaperID{12483}

\begin{document}

\title{Locally Stylized Neural Radiance Fields}

\makeatletter
\renewcommand\AB@affilsepx{ \protect\Affilfont}
\makeatother

\author[1]{Hong-Wing Pang}
\author[2,3]{Binh-Son Hua}
\author[1]{Sai-Kit Yeung}

\affil[1]{Hong Kong University of Science and Technology\hspace{0.2cm}}
\affil[2]{Trinity College Dublin\hspace{0.2cm}}
\affil[3]{VinAI Research, Vietnam}

\maketitle

\begin{abstract}
In recent years, there has been increasing interest in applying stylization on 3D scenes from a reference style image, in particular onto neural radiance fields (NeRF). While performing stylization directly on NeRF guarantees appearance consistency over arbitrary novel views, it is a challenging problem to guide the transfer of patterns from the style image onto different parts of the NeRF scene. In this work, we propose a stylization framework for NeRF based on local style transfer. In particular, we use a hash-grid encoding to learn the embedding of the appearance and geometry components, and show that the mapping defined by the hash table allows us to control the stylization to a certain extent. Stylization is then achieved by optimizing the appearance branch while keeping the geometry branch fixed. To support local style transfer, we propose a new loss function that utilizes a segmentation network and bipartite matching to establish region correspondences between the style image and the content images obtained from volume rendering. Our experiments show that our method yields plausible stylization results with novel view synthesis while having flexible controllability via manipulating and customizing the region correspondences. 

\end{abstract}

\section{Introduction} 
Stylizing a visual world is an increasingly popular and demanding task in games, movies, or extended reality applications. Imagine that one can navigate in an artistic virtual world that resembles the painting styles by different renowned artists. This problem is generally known as 3D style transfer. 

Traditionally, 3D style transfer can be achieved via post-processing. For example, in the well-known traditional computer graphics pipeline, it typically involves a programmable shading stage to post-process the appearance of the rendered geometry or screen images. 
Neural radiance field~\cite{mildenhall2020nerf} is a recent advance in 3D deep learning that aims to represent a 3D scene implicitly by using a neural network trained with multi-view images and differentiable volume rendering. 
Since this pioneer work, significant milestones have been made to greatly improve the performance of neural radiance fields in practice, including improved spatial representation~\cite{liu2020neural}, training convergence~\cite{mueller2022instant}, explicit geometry representation~\cite{wang2021neus}. 
It is therefore promising to revisit the 3D style transfer problem by stylizing a 3D scene implicitly represented by a neural radiance field.

\begin{figure}[t]
\centering
\includegraphics[width=\linewidth]{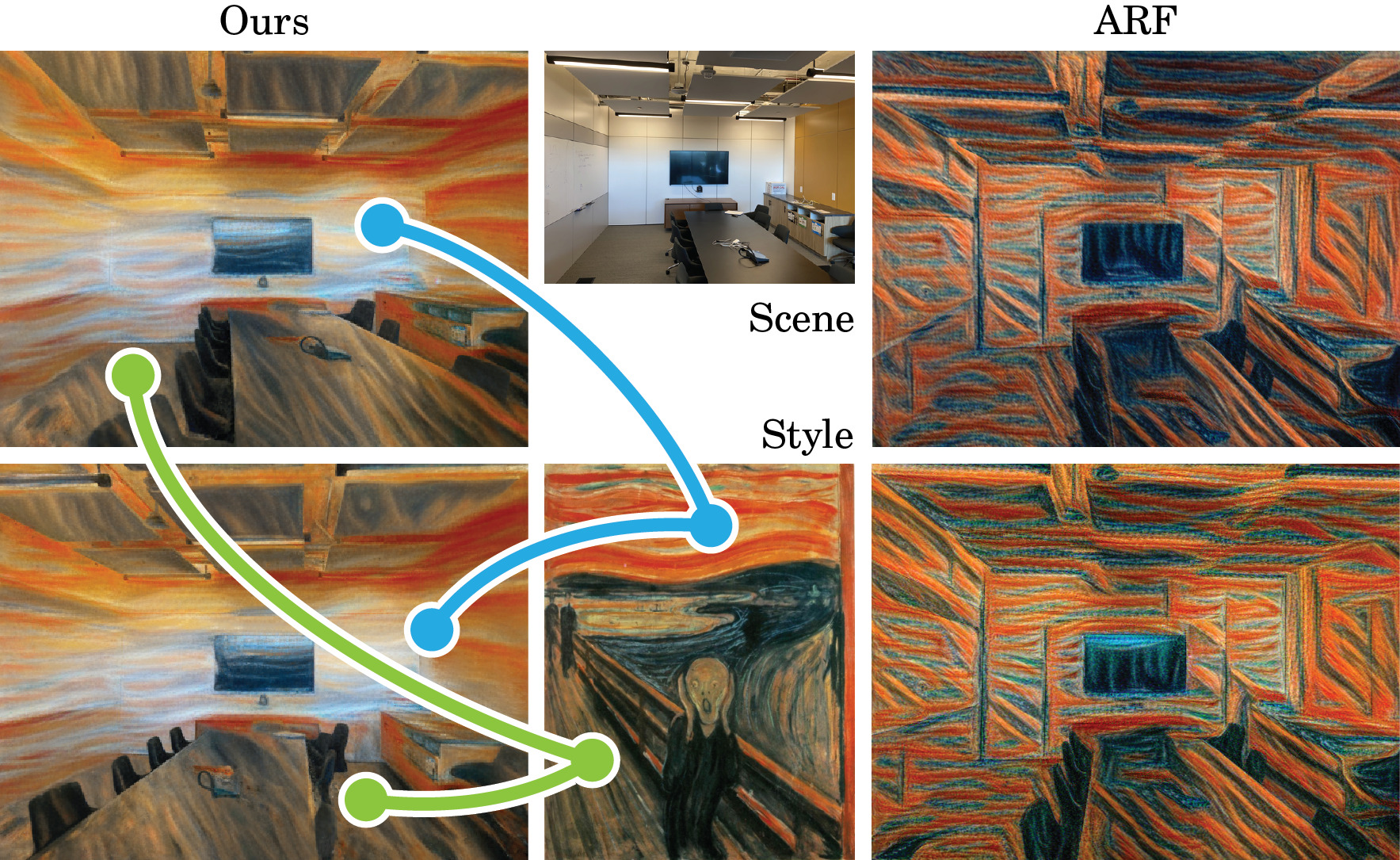}
\caption{We propose a stylization method for NeRF, aware of correspondences between different style patterns and local regions within the rendered image.}
\label{fig:teaser}
\end{figure}

In this work, our goal is to transfer the appearance from a reference style image to a neural radiance field and keep the style transfer consistent across novel views rendered from the radiance field. We are inspired by works in image~\cite{gatys2016image,huang2017adain} and video style transfer methods that have received great attention since the introduction of modern neural networks. While previous works~\cite{nguyen2022snerf,zhang2022arf} have proposed adapting the style transfer problem to neural radiance fields as well, the stylization results lack diversity and controllability.

To address these limitations, we devise a new style transfer method that considers the local transfer between a reference style image and the radiance field rendering. In particular, our method treats style transfer as a post-processing step after the original geometry and appearance of the neural radiance field is learned and therefore aim to perform style transfer while keeping the geometry implicitly represented by the neural radiance field unchanged.
We propose a new backbone suitable for stylizing neural radiance fields that has a dual-branch architecture to learn the density and the appearance field, respectively. We devise a hash-grid encoding scheme with an extended hash function to support storing multiple styles in a single parametric embedding of the appearance field. Given the same reference style image, it is possible to diversify the stylization results by customizing the hash function used for positional encoding.

We further propose a new segmentation-based stylization loss which subdivides both the 3D scene and style image into subregions; different regions in the scene are matched with a region in the style image and stylized accordingly. The matching between scene and style image regions are formulated as a bipartite matching problem and solved by the Hungarian algorithm. We show that our method automatically generates plausible stylization results with high-quality geometry and appearance, reflecting a diverse range of local styles found in the style image. In addition, the generated matching can also be manually edited by the user, making the stylization process controllable.

To summarize, our contributions are: 
\begin{itemize}
    \item A reference-guided style transfer method for neural radiance fields. Our architecture for learning a neural radiance field is a dual-branch network that aims at optimizing the appearance while keeping the geometry fixed during stylization;
    \item An extended hash-encoding scheme for stylization. We provide an analysis of hash-encoding and its influence on the style transfer on radiance fields and multiple style support; 
    \item A new style loss that adopts optimal assignment from bipartite matching on segmented regions between the reference style image and the radiance field rendering for style transfer.
\end{itemize}

\section{Related Work}

\noindent\textbf{Image and video style transfers.}
Image style transfer can be dated back to early work for image analogies~\cite{hertzmann2001analogies}, which uses a pair of images as the training data to learn a filter which can be subsequently applied on new images to simulate analogous filtered results. 
Deep image analogies~\cite{liao2017deepanalogies} performs visual attribute transfers by using deep features from a neural network to derive dense correspondences between the input and the style image, which is only applicable when these images are semantically similar, e.g., images of human faces. 
Neural style transfer~\cite{gatys2016image} instead uses deep features to build content and style constraints and optimizes a random noise image to output a new image with matched content and style statistics. 
The optimization can be replaced by a feed-forward prediction by training a neural network on pairs of original and stylized images with improved results using instance normalization~\cite{ulyanov2016instancenorm} and perceptual loss~\cite{johnson2016perceptual}. 
Recent methods aim at supporting arbitrary style images at test time without the need to retrain the stylization network by leveraging adaptive instance normalization~\cite{huang2017adain}, patch-based transfer~\cite{texler2020patch,texler2019patch}, feature transform~\cite{li2017universal} as well as multiple style support~\cite{sanakoyeu2018styleaware}. 

As an extension of image style transfer methods, video style transfer methods further deal with the temporal consistency problem for transferring styles across video frames~\cite{jamriska2019videobyexample,texler2020interactivestyle}. 
Different from video style transfer methods, we focus on stylization of 3D scenes, where view-consistent stylization is required to achieve high-quality novel view synthesis. 

\begin{figure*}[h]
\centering
\includegraphics[width=\textwidth]{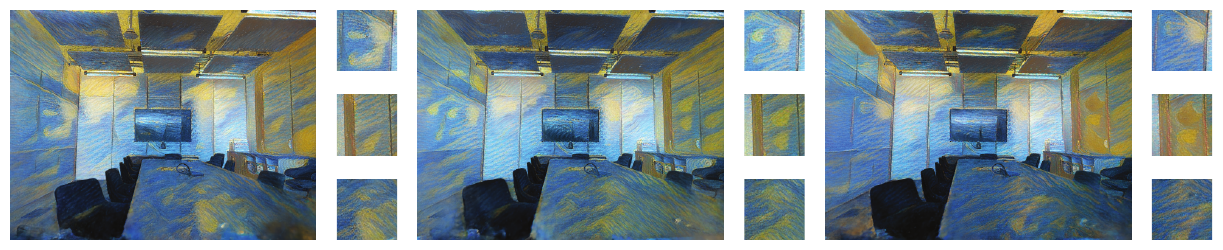}
\caption{Stylization results from modifying the hash function coefficients. This yields similar and consistent global styles while having local diversity.}
\label{fig:hash}
\end{figure*}

\noindent\textbf{3D style transfers.} 
Stylizing a 3D scene can be performed explicitly on point clouds and mesh representations~\cite{kato2018renderer,huang2021iccv,hollein2022stylemesh}. 
However, this approach is error-prone due to imperfect geometry and texture rendering. 
In contrast, NeRF~\cite{mildenhall2020nerf} allows representing a 3D scene by learning a neural network using multiple-view images and differentiable volume rendering. As a result, NeRF can smoothly and consistently interpolate its rendering at different angles, which is an ideal application for novel view synthesis.
The last few years witness tremendous research interests in this area with significant progress in improving NeRF in various aspects including its spatial representation~\cite{liu2020neural}, training convergence~\cite{mueller2022instant}, explicit geometry representation~\cite{wang2021neus}, to name a few. 
Stylizing a 3D scene by using a NeRF representation is therefore gaining attention recently. 

A simple approach to stylize a NeRF model is to regress a NeRF and constrain its rendering to 2D image stylization results with a content loss and a style loss. 
SNeRF~\cite{nguyen2022snerf} follows this approach and demonstrate consistent stylization across novel views. 
Their style loss is a global constraint and ignores any spatial correspondences between the rendering and the style image. 
ARF~\cite{zhang2022arf} defines a nearest-neighbor feature matching loss but since their correspondence assignment is fixed, their results do not support diverse stylization results. 
Our work differs in that we propose a spatial matching between the style image and NeRF rendering so that the dynamically assigned correspondences can be used to guide the stylization with diverse results. 

To speed up stylization process, the optimization on NeRF can be replaced by a feed-forward prediction. 
Chiang et al.~\cite{chiang2022hypernetwork} used a hypernetwork to predict the weights of a NeRF MLP given an arbitrary style. 
Our work utilizes the hash encoding~\cite{mueller2022instant} for style transfer, which significantly speeds up the optimization process.

Other attempts have been made to improve consistency in the 2D and 3D space during style transfer such as using a point cloud as an intermediate representation for 2D-3D feature transfer~\cite{huang2021iccv}, or using mutual learning by distilling spatial consistency and stylized rendering between a NeRF and a 2D style transfer network~\cite{huang2022stylizednerf}. 

\section{Our Method} 

The stylization of NeRF models are typically completed in two stages - the \textit{reconstruction stage}, where a base NeRF model is trained with respect to the MSE reconstruction loss, similar to regular NeRF training; followed by the \textit{stylization stage}, where the parameters of the NeRF model is fined tuned against style-transfer specific losses. Our method follows this paradigm, but we reconsider several design choices discussed below.

\subsection{Dual-branch NeRF model}

\begin{figure}[t]
\centering
\includegraphics[width=0.48\textwidth]{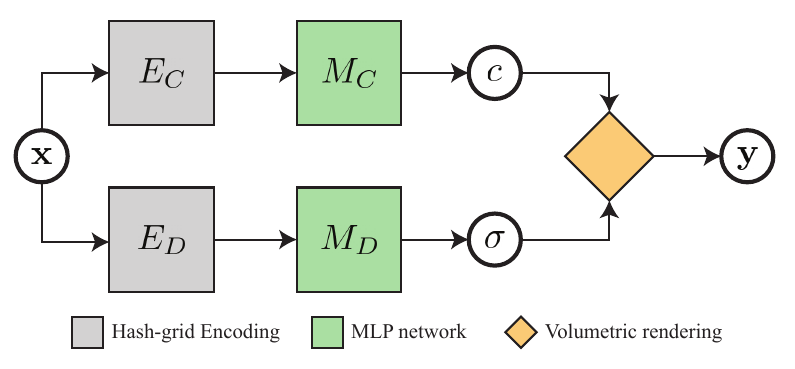}
\caption{Our dual-branch architecture for neural radiance fields. We optimize only the appearance branch for the stylization task.}
\label{fig:ngp_diagram}
\end{figure}

In this work, we propose a dual-branch architecture for our neural radiance fields. Our model is illustrated in Figure \ref{fig:ngp_diagram}. Our dual-branch architecture consists of a geometry branch to encode the density component of the neural radiance field, and an appearance branch to encode the color component, respectively. This design allows us to subsequently only optimize the appearance branch for the stylization task while keeping the geometry unchanged. 

We utilize contributions from Instant-NGP~\cite{mueller2022instant} to represent each branch, which uses a parametric embedding to replace the original fixed positional embedding used in vanilla NeRF. 
Particularly, the bounded scene volume is subdivided into a large number of voxels each corresponding to a set of learnable parameters, which is stored in a fixed-size hash map. Thus, the learned embedding provides a rich, descriptive set of features representing the scene geometry and appearance. The depth and no. of parameters of the subsequent MLP networks can be greatly reduced, improving the training time by a few orders of magnitude.
The proposed model differs from the vanilla architecture proposed in Instant-NGP~\cite{mueller2022instant} in the following ways. First, we train two separate sets of hash-grid encodings, $E_C$ and $E_D$ to represent the appearance and geometry, respectively. Second, we discard the view direction input. This is a common practice found in prior NeRF stylization work, considering that style-transferred scenes often do not require view-dependent effects during rendering.

During the reconstruction stage, we train our model with multi-view images and the MSE loss, similar to ~\cite{mueller2022instant}. During the stylization stage, we fix the weights of components $E_D$ and $M_D$, such that the density value $\sigma$ corresponding to each input point $\mathbf{x}$ remains fixed. We then optimize the appearance component $E_C$ and $M_C$ with a novel style loss, to be discussed in Section \ref{section:segmentation}.

\paragraph{Effect of hash-grid encoding in stylization.} In our formulation, the large number of parameters in the learned hash-grid encoding $E_C$ describes the rendered appearance of the radiance field. Specifically, the encoding corresponding to the voxel at $(x,y,z)$ is given by the geometric hash function $H$ described in ~\cite{mueller2022instant}:
\begin{align}
H(x,y,z)=(h_1x\oplus h_2y\oplus h_3z)\mod N_H,
\end{align}
where $h_1,h_2,h_3$ are large prime numbers and $N_H$ is the total size of the hash-grid. By picking different values for $h_1,h_2,h_3$, we demonstrate in Figure \ref{fig:hash} that different patterns can be generated after the stylization training stage. Compared with existing stylization methods where a single stylization result is generated per style image / scene, this gives a wider level of variation.

In addition, our architecture can be used to stylize multiple styles simultaneously, as opposed to prior methods that focus on transferring a single style. This is done by adding a fourth term to the hash function:
\begin{align} 
H(x,y,z)=(h_1x\oplus h_2y\oplus h_3z\oplus h_4s_i)\mod N_H,
\end{align}
where $s_i$ is a discrete style index. In other words, we store encodings for multiple styles inside the same hash-grid, allowing stylization of more than one style at inference time. We demonstrate the results of training multiple styles simultaneously in Section 3 of the supplementary material.

\subsection{Preliminaries on style loss}

Given a pair of rendered output image $\mathbf{y}$ and style image $\mathbf{s}$, stylization losses typically operate on high-level features $\mathbf{f}_y=\mathcal{F}(\mathbf{y}),\mathbf{f}_s=\mathcal{F}(\mathbf{s})$ extracted with a pre-trained and fixed feature extraction network $\mathcal{F}$; for example, StylizedNeRF~\cite{huang2022stylizednerf} follows the training setup of AdaIN~\cite{huang2017adain}, which uses a style loss that matches statistics (i.e. mean and standard deviation) between $\mathbf{f}_y$ and $\mathbf{f}_s$, where $\mathcal{F}$ is a pretrained VGG-16~\cite{simonyan2015vgg} network.

However, the matching of global statistics means that the style may not be properly transferred across every local region in $I$. The nearest-neighbor feature matching (NNFM) loss introduced in ~\cite{zhang2022arf} uses the following formulation instead:
\begin{align}
\mathcal{L}_{\text{NNFM}}(\mathbf{y},\mathbf{s})=\frac{1}{N_F}\sum_{f_i\in\mathbf{f}_y}\min_{f_j\in\mathbf{f}_s}d(f_i,f_j)
\end{align}
where every individual feature vector $f_i$ in $\mathbf{f}_y$ is paired with the closest style feature vector $f_j$ in $\mathbf{f}_s$, in terms of the cosine angle distance $d$. The NNFM loss is then defined as the mean of distances over all $N_F$ pairs of vectors, where $N_F$ is the no. of vectors in $\mathbf{f}_y$.

\begin{figure}[t]
\centering
\includegraphics[width=0.48\textwidth]{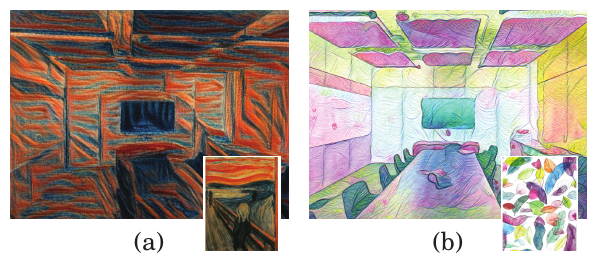}
\caption{Stylization results using nearest-neighbor feature matching (NNFM)~\cite{zhang2022arf} are not always satisfactory, for example, recurrent patterns occur in the stylization results and not all regional styles are transferred. }
\label{fig:arf}
\end{figure}

However, matching up nearest neighbor features between $\mathbf{f}_y$ and $\mathbf{f}_s$ may not always lead to good results. Figure \ref{fig:arf} shows two stylization results by ARF onto the same scene. We see that in both cases, recurrent patterns are generated over the entire scene. In (a), different parts of the scene (e.g. floor, walls) gets the same pattern, even though there are multiple distinct patterns in the original style image to choose from. In (b), recurrent patterns are generated on the ``blank" walls, even when equally ``blank" regions are present in the style image. We suggest that applying nearest neighbor search over the entire image does not lead to the best choice in stylization, especially in cases where vectors in $\mathbf{f}_y$ are dissimilar to all vectors in $\mathbf{f}_s$.

\subsection{Style loss with region correspondences}
\label{section:segmentation}

\begin{figure*}[h]
\centering
\includegraphics[width=\textwidth]{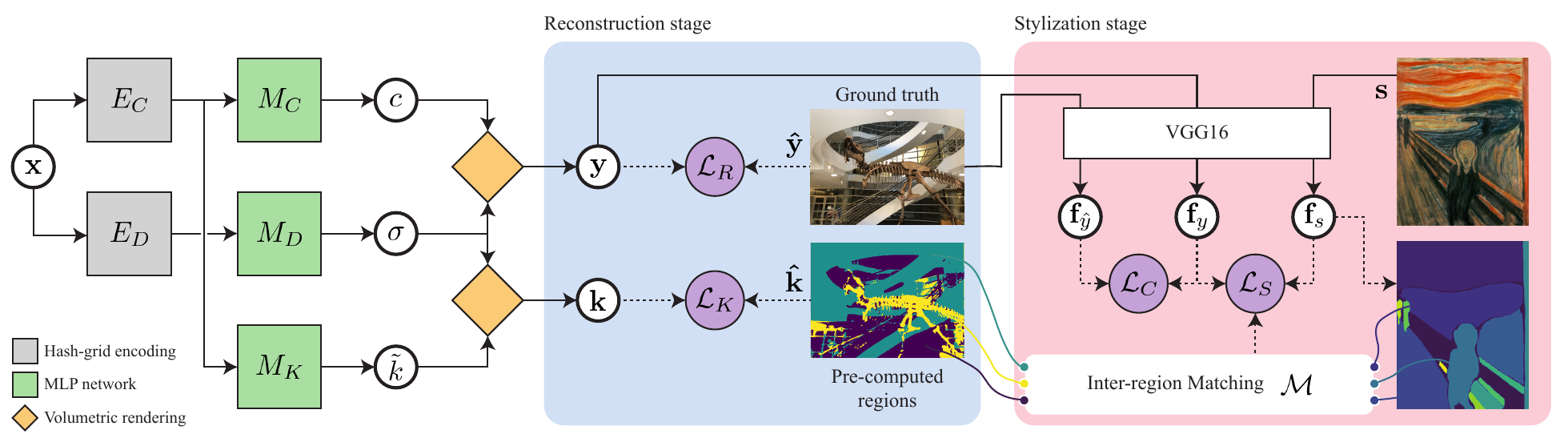}
\caption{Our two-stage pipeline for NeRF stylization. During the reconstruction stage, the entire network is trained to simultaneously render a novel view $\mathbf{y}$ of the target scene and generate a segmentation map $\mathbf{k}$. During the stylization stage, the appearance components ($E_C$ and $M_C$) are fine-tuned with our novel style loss $\mathcal{L}_S$ which considers the matching between regions in $\mathbf{y}$ and $\mathbf{s}$.}
\label{fig:stylization_diagram}
\end{figure*}

Our key observation is that, instead of only comparing between nearest neighbors on the feature level, we can divide up $\mathbf{y}$ and $\mathbf{s}$ into coarse regions $\{\mathbf{y}_i\}$ and $\{\mathbf{s}_j\}$, where features in region $\mathbf{y}_i$ is matched towards features within a corresponding region $\mathbf{s}_j$. The idea of matching feature statistics between regions has previously been explored in \cite{luan2017deep}, which segments content and style images into regions by semantic segmentation and pair-up the regions by semantic labels. It is nontrivial, however, to extend this method to NeRF stylization, for the following reasons. 
First, in order to stylize an entire scene, the scene will need to be dissected into regions in a manner that is consistent over any arbitrary novel view. 
Second, the method proposed in \cite{luan2017deep} is designed for photorealistic style images; in contrast, semantic segmentation is unlikely to give meaningful results on artistic style images.
Third, the lack of semantic segmentation labels also means that there is no intuitive way to pair-up the regions.

To this end, we propose a training pipeline which segments \textit{both} $\mathbf{y}$ abd $\mathbf{s}$. A matching is derived automatically from the segmented regions, which is used to influence the calculation of the style loss. An overview of this pipeline is described in Figure \ref{fig:stylization_diagram}.

\paragraph{Defining regions for $\mathbf{y}$ and $\mathbf{s}$.}
The first step to our pipeline is to subdivide $\mathbf{y}$ and $\mathbf{s}$ into $C$ \textit{scene regions} and $S$ \textit{style regions}. We use the unsupervised segmentation method \textit{Segment Anything}~\cite{kirillov2023segany} to segment $\mathbf{s}$ into $S$ style regions with diverse local patterns and styles.

Computing the segmentation of $\mathbf{y}$ is nontrivial, as we require a consistent segmentation that assigns regions consistently over any arbitrary novel views from the same scene. To solve this problem, we first segment the set of training images $\{\mathbf{\hat{y}}\}$ into $C$ regions, using a variant of the \textit{unsupervised} image segmentation method proposed by Kim et. al.~\cite{kim2020segment}; which allows simultaneous segmentation of multiple unseen images without prior semantic knowledge.

We then introduce an additional MLP network $M_K$ in our NeRF backbone as shown in Figure \ref{fig:stylization_diagram}, which produces a $C$-dimensional vector output:
\begin{align}
\tilde{k}=M_K(E_c(x)).
\end{align}

The list of $\tilde{k}$s computed over a single ray are integrated together with density values from $M_D$ using the volumetric equation, and subsequently passed through the softmax function. The result is a probability vector $k\in\mathbb{R}^C$ describing the probabilities of the pixel belonging to each of the $C$ scene regions. During the reconstruction stage, $M_K$ is trained simultaneously with other components with respect to the cross-entropy loss $\mathcal{L}_K$. For each vector $k_i$ computed from a pixel, we have
\begin{align}
\mathcal{L}_K(k_i,\hat{k}_i)=\sum_{i=1}^C-\hat{k}_ik_i\log k_i,
\end{align}
where $\hat{k}_i$ is the corresponding scene region (in one-hot vector form) from the segmentation map $\mathbf{\hat{k}}$, pre-computed from the ground truth training image $\mathbf{\hat{y}}$. After the reconstruction stage is completed, the NeRF backbone can simultaneously render a novel view $\mathbf{y}$ and generate its corresponding segmentation map $\mathbf{k}$. Note that our stylization pipeline is agnostic to the segmentation method used to segment $\mathbf{s}$ and $\{\mathbf{\hat{y}}\}$. The choice of $C$ and $S$ are automatically determined during unsupervised segmentation; for the results shown on this paper, we have $C\approx5$ and $S\approx10$. 

In Section 1 of the supplementary material, we describe the implementation of segmentation in greater detail, as well as provide an ablation experiment to demonstrate the effect of using higher values of $C$ through segmenting the scene and style image into finer regions.

\paragraph{Style loss.}
Following the procedure in Section \ref{section:segmentation}, we have $C$ scene regions $\{\mathbf{y}_i\}$ segmented from $\mathbf{y}$; as well as $S$ style regions $\{\mathbf{s}_j\}$ segmented from $\mathbf{s}$. To avoid multiple regions being mapped to a single local pattern in $\mathbf{s}$, we formulate this as a bipartite matching problem where no two scene regions are mapped to the same style region.

To do so, we construct a cost matrix $\mathbf{W}\in\mathbb{R}^{C\times S}$, where each individual entry $W_{ij}$ represents the affinity between regions $\mathbf{y}_i$ and $\mathbf{s}_j$. The cost is determined by the \emph{feature distance} and the \emph{patch distance}, defined as follows. The feature distance is defined as the cosine angle distance between the means of features in $\mathbf{y}_i$ and $\mathbf{s}_j$. The patch distance is defined as the Euclidean distance between the centroids (i.e. arithmetic mean position of all pixels constituting the region) of $\mathbf{y}_i$ and $\mathbf{s}_j$; each centroid position is normalized as a value in $[0,1]$ given the image dimensions.
The patch distance dictates that scene regions are more likely to be paired with style regions that are roughly located in the same relative position. Using the previous example in Figure \ref{fig:arf} (a), the scene regions corresponding to the ceiling and floor are more likely to be mapped to patches in the sky and floor in the style image.

After computing $\mathbf{W}$, an optimal injective mapping $\mathcal{M}:[1,C]\mapsto[1,S]$ can be obtained by applying the Hungarian algorithm, such that the following mapping cost $\sum_{i\in C}W_{i,\mathcal{M}(i)}$ is minimized. Given a mapping $\mathcal{M}$, our updated style loss is formulated as follows:
\begin{align}
\mathcal{L}_S(\mathbf{y},\mathbf{s},\mathcal{M})=\frac{1}{N_F}\sum_{f_i\in\mathbf{f}_y}\min_{f_j\in\mathbf{s}_j}d(f_i,f_j)
\end{align}
such that if $f_i\in\mathbf{y}_i$, then $\mathbf{s}_j$ is the corresponding style region of $\mathbf{y}_i$ in $\mathcal{M}$.

\paragraph{Custom matching.} While the above method provides an automatic matching between $\{\mathbf{y}_i\}$ and $\{\mathbf{s}_j\}$, the exact pairing can be further modified to produce a wide variety of style transfer results for different scenarios. We demonstrate this capability in Section \ref{section:different_pairs}.

\begin{figure*}[h!]
\centering
\includegraphics[width=\textwidth]{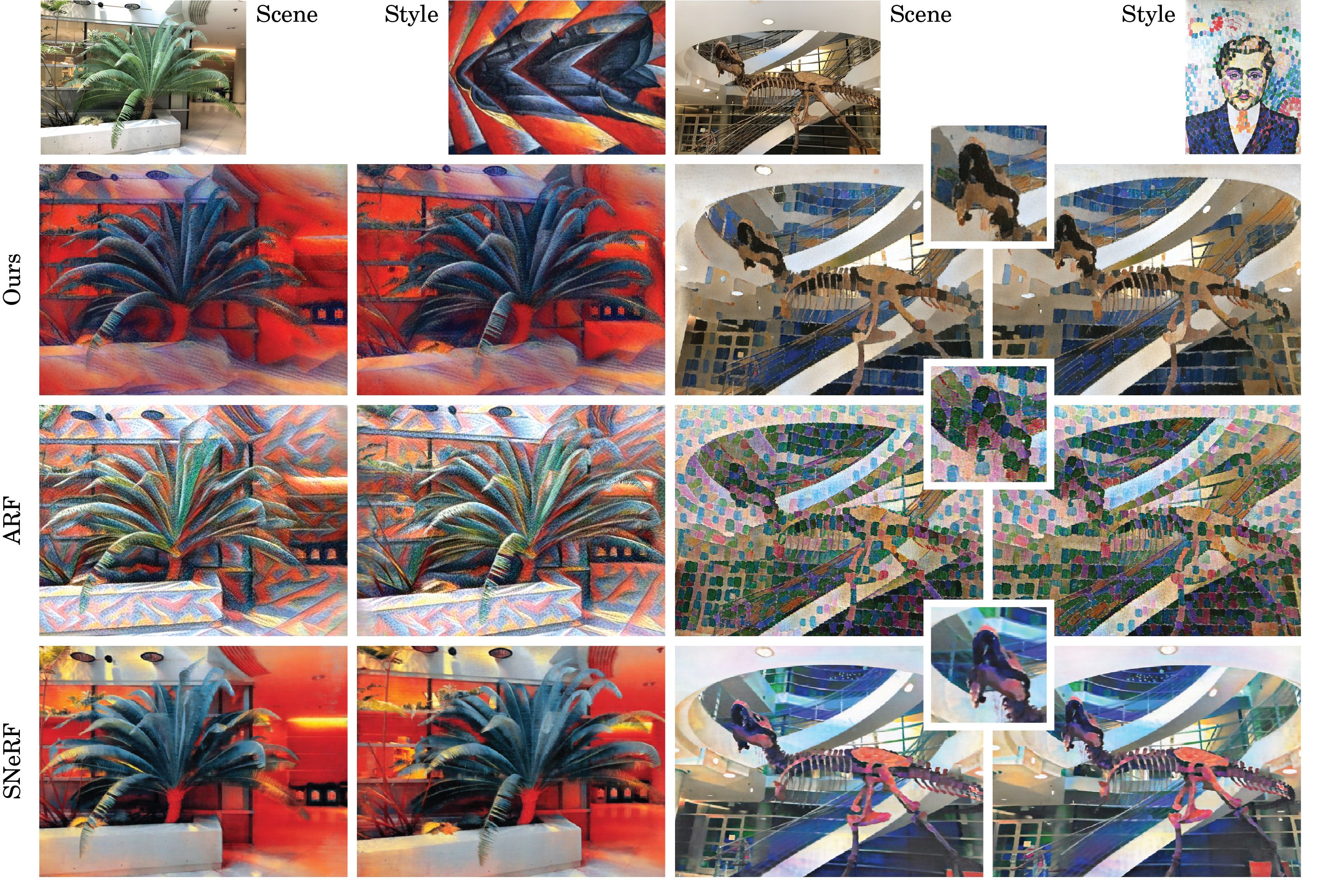}
\caption{Qualitative comparison results with SNerf~\cite{nguyen2022snerf} and ARF~\cite{zhang2022arf} on the LLFF dataset~\cite{mildenhall2019llff}.}
\label{fig:snerf_comparison}
\end{figure*}

\subsection{Training and implementation details}
\label{section:training}

\paragraph{Reconstruction stage.} The model is first trained for 20,000 iterations with the objective function:
\begin{align}
\mathcal{L}_R(\mathbf{y}, \mathbf{\hat{y}})+\lambda_{CE}\sum_{i=1}^{N_T}\mathcal{L}_K(k_i, \hat{k}_i),
\end{align}
where we mix the reconstruction loss $\mathcal{L}_R$ with the cross-entropy loss $\mathcal{L}_K$ multiplied by a fixed constant $\lambda_{CE}=0.01$. $N_T$ is the number of pixels sampled during each training iteration. It is further trained for 2,000 iterations after applying color transformation from $\mathbf{s}$ to $\mathbf{y}$, as proposed in \cite{zhang2022arf}.

\paragraph{Stylization stage.} The pairing $\mathcal{M}$ between scene regions and style regions is first computed using the heuristic functions described above. We fix $\mathcal{M}$ through the entire stylization stage to avoid instability during training. We then train the model with the following objective function:
\begin{align}
\lambda_{C}\mathcal{L}_C(\mathbf{y}, \mathbf{\hat{y}})+\lambda_{S}\mathcal{L}_S(\mathbf{y}, \mathbf{s}, \mathcal{M}),
\end{align}
where $\mathcal{L_C}$ is the content loss $\|\mathbf{f}_y - \mathcal{F}(\mathbf{\hat{y}})\|_2^2$; $\lambda_{C}$ is fixed to 0.001; and $\mathcal{L_S}$ is the style loss with the pairing $\mathcal{M}$ enforced.

In addition, we follow the experimental settings in \cite{zhang2022arf}, where the feature extractor $\mathcal{F}$ collects all the post-ReLU features from the \texttt{conv3} block of VGG-16 and concatenates them together; as well as utilizing \textit{deferred backpropagation} to allow optimizing the loss function over entire images under limited GPU memory.

\section{Experiments}

\subsection{Qualitative comparisons}
We demonstrate qualitative comparisons in Figure \ref{fig:snerf_comparison} with ARF~\cite{zhang2022arf} and SNerf~\cite{nguyen2022snerf}, another relatively recent baseline for NeRF stylization. As SNerf do not release their code implementation, we perform comparison on the same scenes and style images used in Figure 5 in their paper, using scenes in the LLFF~\cite{mildenhall2019llff} dataset.

\begin{figure*}[h]
\centering
\includegraphics[width=\textwidth]{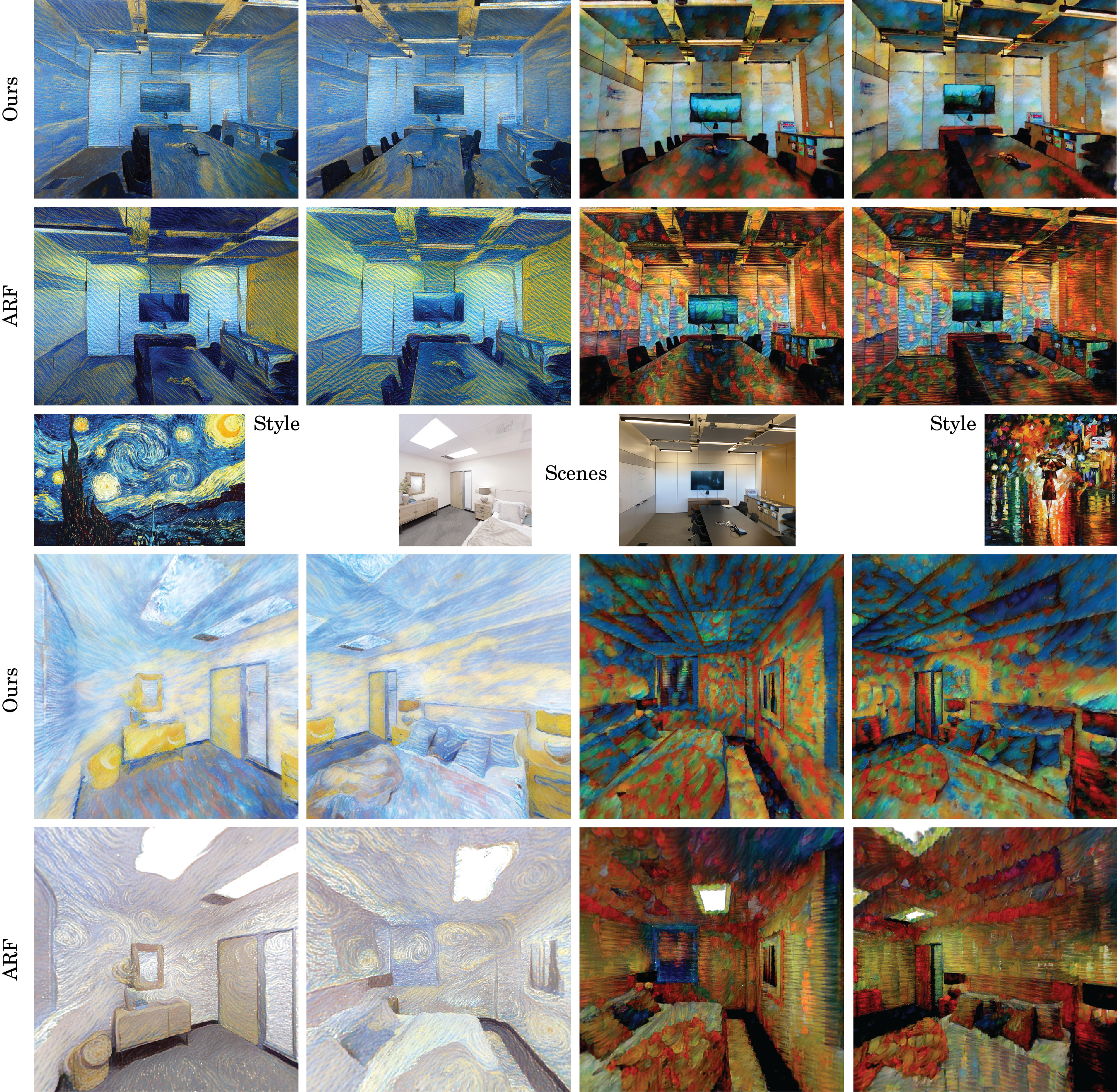}
\caption{Further qualitative comparison results with ARF~\cite{zhang2022arf} on an LLFF scene~\cite{mildenhall2019llff} (top rows) and on a Replica scene~\cite{replica19arxiv} (last rows).}
\label{fig:arf_comparison}
\end{figure*}

In general, the results produced by SNerf do not have significant change in terms of the pattern of the image; only the color has been altered to match the style image. SNerf directly computes the MSE loss between $\mathbf{f}_y,\mathbf{f}_s$ as their style loss, meaning that the style similarity is being compared on an image-wide level. Thus, it is hard to predict which style region will be transferred to which scene region. ARF is successful in transferring local patterns from the style image; however, in either scenes large parts of the scene has been blanketed with the same pattern. In the second scene, the dinosaur head is difficult to identify after applying the mosaic pattern.

Our results strike a balance between the two baselines. In the second scene of Figure~\ref{fig:snerf_comparison}, the background area is mapped to the dark blue suit in the style image; whereas the top area is mapped to the top of the style image where the mosaic pattern is not as obvious. The matching is both consistent and predictable, resulting in a diverse combination of patterns over the entire scene.

We further compare our stylization results with ARF on more examples in Figure \ref{fig:arf_comparison}. We use another scene in the LLFF dataset, followed by another scene in the Replica~\cite{replica19arxiv} dataset. Our method is capable of drawing diverse patterns and styles within the given style image, applying to consistent regions across the content image. For example, in the style image to the right, we can better distinguish different parts of the scene by the use of different patterns and colors. We provide further qualitative comparisons of even more scenes and style images in the supplementary material.

\begin{figure*}[h]
\centering
\includegraphics[width=\linewidth]{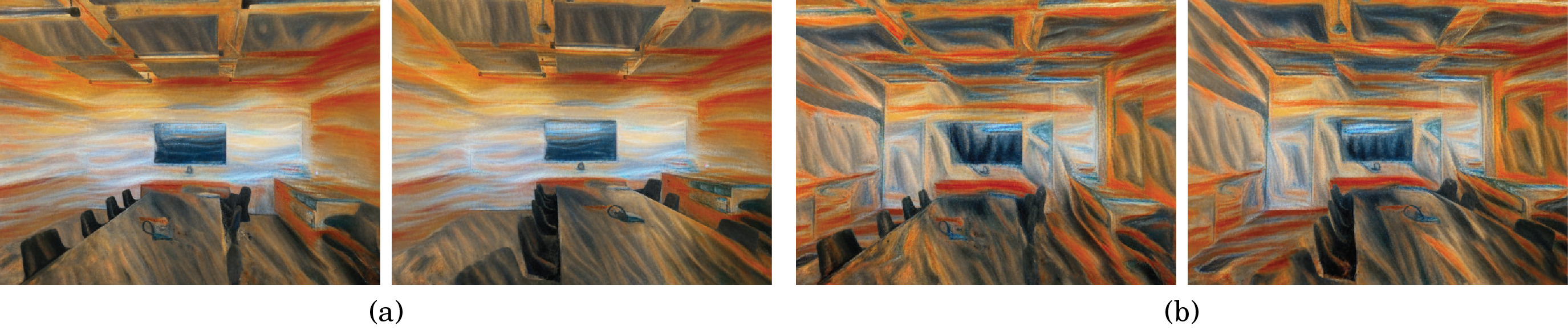}
\caption{Ablation comparison with (a) and without (b) pairing between scene and style regions.}
\label{fig:nnfm_ablation}
\end{figure*}

\begin{figure*}[h]
\centering
\includegraphics[width=\linewidth]{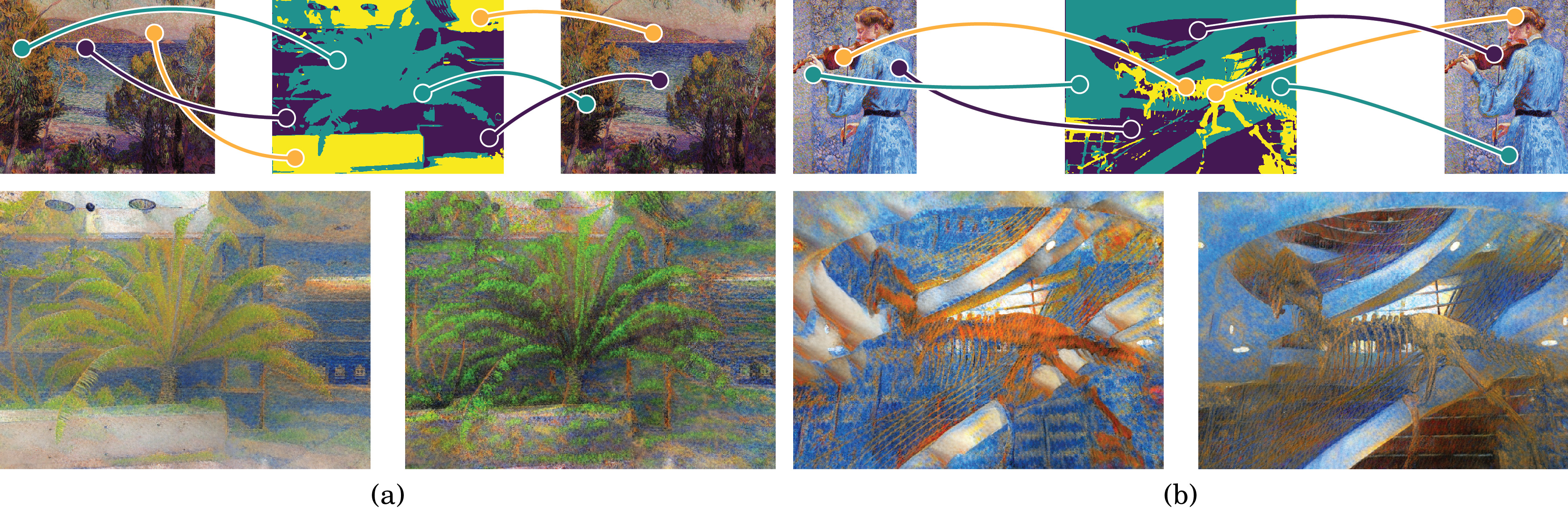}
\caption{Stylization results via custom pairings between scene and style regions. Best view with zoom.}
\label{fig:different_pairs}
\end{figure*}

\subsection{Ablation experiments}
It is mentioned previously in Section \ref{section:training} that the pairing $\mathcal{M}$ is determined at the start of the stylization stage, and influences the computation of style loss $\mathcal{L}_S$. Without this pairing, the style loss falls back to the NNFM loss which looks for the nearest neighbor feature across the entire style image.
In Figure \ref{fig:nnfm_ablation}, we evaluate the effect of computing $\mathcal{M}$ by comparing the results with using the vanilla NNFM loss on our NGP-based rendering backbone.

\subsection{Effect of different pairings}
\label{section:different_pairs}

In Figure \ref{fig:different_pairs}, we demonstrate the capability to customize the stylization result via modifying the pairing $\mathcal{M}$, given the same scene and style image. In both cases, the scene is segmented into 3 classes, which are manually mapped to two different set of style regions. In (a), the tree in the foreground undergoes a significant change in appearance when it is mapped to different plants in the style image; whereas in (b), the scene regions also changes appearance based on the different pairing.

\section{Further discussion}

\noindent\textbf{Extension to unbounded scenes.} While Instant-NGP provides a form of 3D scene representation that is efficient to render, it requires the scene to be contained within a bounded volume, which will be mapped to the 3D hash grid for parameter storage. This makes it non-trivial to extend our method towards unbounded scenery, e.g. Tanks \& Temples~\cite{Knapitsch2017}. It would be helpful to extend the hash-grid encoding towards unbounded scenes by reparametrizing the coordinates within a bounded volume, such as the inverse-distance approach used in NeRF++~\cite{kaizhang2020}.

\noindent\textbf{Optimizing density.} In contrast to stylization methods where the density component of the backbone is fixed, recent methods such as SNerf~\cite{nguyen2022snerf} and NerfArt~\cite{wang2022nerf}) has suggested that optimizing both the density and appearance components will lead to improvement in stylization results. It is debatable whether optimizing the density field is really necessary or not. 
We opt for leaving density untouched. 

\section{Conclusion}

In this work we present a novel framework for stylizing a neural radiance field. We optimize the appearance branch in a dual-branch radiance field represented by a parametric embedding learned with an extended hash function. We devise a new style loss based on region correspondences between the style image and the content images from rendering the radiance field. Our method generates diverse stylization results, where the stylization can be controlled by the extended hash function and the region correspondences. 

Our method is not without limitations. First, our method is not (yet) suitable for interactive use even though the hash-encoding scheme reduces the required time for optimizing the neural radiance fields to minutes. Second, our method support multiple styles but these styles have to be known in advance of the stylization. Supporting arbitrary style transfer at test time would be an interesting future work. 

\noindent\textbf{Acknowledgment.} This paper was partially supported by an internal grant from HKUST (R9429) and the HKUST-WeBank Joint Lab.

\appendix

\section*{Supplementary material}

In this supplemental document, we provide additional details and experiment results for our proposed method. In particular, we provide an in-depth description of the segmentation procedure used in our stylization procedure; a user study; and more qualitative results on the LLFF and the Replica dataset.

\section{Segmentation procedure}

In this section, we describe in greater detail the segmentation procedures used for creating scene regions and style regions. As mentioned in the main paper, our stylization method can be performed irrespective of the segmentation method used.

\subsection{Scene regions}
\label{section:supp_scene_regions}

Given a set of training images $\{\mathbf{\hat{y}}\}$, we want to generate a set of segmentation maps $\{\mathbf{\hat{k}}\}$, where every pixel is classified into a finite set of $C$ scene regions. The segmentation process should be unsupervised (i.e. $C$ is not set to a fixed value, and is determined during segmentation), and can be applied to any arbitrary set of scenes.

To satisfy these requirements, we base our implementation on the segmentation procedure proposed by Kim et. al. \cite{kim2020segment}. In their method, a single image is passed into a CNN model producing a \textit{response map} $\mathbf{r}\in\mathbb{R}^{H\times W\times Q}$, where $Q$ is the upper bound of no. of scene regions. A segmentation map $\mathbf{c}\in\mathbb{R}^{H\times W}$ can be obtained from $\mathbf{r}$ by taking the argmax function.

This network is trained with a combination of the \textit{similarity loss} $\mathcal{L}_{sim}$ and \textit{continuity loss} $\mathcal{L}_{con}$. The similarity loss is defined as the sum of cross-entropies between response vector $r_n$ and the target vector $c_n$:
\begin{align}
	\mathcal{L}_{sim}(\mathbf{r})=-\sum_{n}\sum_{i=1}^Qc_{n,i}\log{r_{n,i}},
\end{align}
where $c_n$ is the one-hot vector in $\mathbf{c}$ corresponding to $r_n$.
The continuity loss is defined as the sum of L1 distances between horizontally and vertically adjacent features in $\mathbf{r}$:
\begin{align}
	\mathcal{L}_{con}(\mathbf{r})=\sum_{i=1}^{W-1}\sum_{j=1}^{H-1}\|r_{i+1,j}-r_{i,j}\|_1+\|r_{i,j+1}-r_{i,j}\|_1.
\end{align}
It can be seen that $\mathcal{L}_{sim}$ encourages similar features in $\mathbf{r}$ to be grouped together and form a single cluster; $\mathcal{L}_{con}$ ensures spatial continuity of clusters and prevents the segmentation from being too fragmented. In general, the unique number of classes in $\mathbf{c}$ is initially high (i.e. close to $Q$), and gradually decreases over time as more and more feature vectors in $\mathbf{r}$ are clustered into the same region.

To extend this method to segmentation of multiple images simultaneously, we train the segmentation network by sampling a batch of $B$ images during each iteration, instead of a single image. The loss values for each individual response maps $\{\mathbf{r}_1,\cdots,\mathbf{r}_B\}$ are computed and summed together. After the network is trained, we can run segmentation over all of $\{\mathbf{\hat{y}}\}$, obtain the set of $C$ remaining active classes, and re-index the segmentation maps from 1 to $C$.

\subsection{Style regions}

Given a style image $\mathbf{s}$, we want to segment it into a set of $S$ style regions $\{\mathbf{s}_j\}$, once again without explicit supervision. Unlike section \ref{section:supp_scene_regions}, we only need to apply segmentation on a single image. We use the robust \textit{Segment-Anything} method \cite{kirillov2023segany} as it has good performance outside real-life images, which is the case for style images in artistic style transfer. We use the official pretrained weights based on ViT-H~\cite{DBLP:conf/iclr/DosovitskiyB0WZ21}.

Directly applying the method results in a set of regions $\{\mathbf{s}_j\}$ which may overlap with each other. To fix this issue, we first sort the regions in decreasing order of size, and run the procedure in Algorithm~\ref{alg:cap}.
\begin{algorithm}
	\caption{Filtering overlapping style regions}\label{alg:cap}
	\begin{algorithmic}
		\State $\mathbf{m} \gets 0$
		\State $\mathbf{k} \gets -1$
		\State $i \gets 0$
		\For{$\mathbf{s}_j$ in $\{\mathbf{s}_1,\cdots,\mathbf{s}_N\}$}
		\If{$\sum\mathbf{m}[\mathbf{s}_j]/|\mathbf{s}_j|\ge\lambda_t$ and $|\mathbf{s}_j|/\|\mathbf{s}\|\ge\lambda_m$}
		\State $(\mathbf{m}[\mathbf{s}_j]) \gets 1$
		\State $(\mathbf{k}[\mathbf{s}_j]) \gets i$
		\State $i \gets i + 1$
		\EndIf
		\EndFor
	\end{algorithmic}
\end{algorithm}
Here, $\{\mathbf{s}_1,\cdots,\mathbf{s}_N\}$ is the list of regions from largest to smallest; $\mathbf{m}\in\mathbb{R}^{H\times W}$ is a binary map that keep tracks if a pixel has been assigned to a style region; and $\mathbf{k}\in\mathbb{R}^{H\times W}$ is the segmentation map. The notation $\mathbf{m}[\mathbf{s}_j]$ and $\mathbf{k}[\mathbf{s}_j]$ represents the subset of pixels in $\mathbf{m}$ and $\mathbf{k}$ belonging to region $\mathbf{s}_j$. $\lambda_t$ determines if the current $\mathbf{s}_j$ is overlapping with previous regions; $\lambda_m$ ensures that regions too small are not considered. We set $\lambda_t$ as 0.05 and $\lambda_m$ as 0.004 (i.e. 0.4\% of total image area).

After the procedure, each pixel of $\mathbf{k}$ should be given an integer value from $[-1,S-1]$, where $0$ to $S-1$ indicates the $S$ style regions. An index of -1 means that the pixel is not assigned to any region, and is not considered during stylization.

Last but not least, $\mathbf{k}$ is downscaled by nearest neighbor interpolation to match the dimensions of $\mathbf{f}_s$, the VGG features extracted from $\mathbf{s}$.

\subsection{Fine vs. coarse regions}

The values of $C$ and $S$ determines the level of fineness during the segmentation of scene and style regions. We provide an ablation experiment to experiment on the effect of using larger values of $C$ and $S$ on the stylization result.

The value of $C$ can be indirectly controlled by modifying the number of iterations of training the unsupervised segmentation network; in general, by using a smaller number of iterations, the network output will contain a larger number of classes as it has not fully converged. For this experiment, we segment the style image $\mathbf{s}$ with the same procedure as scene regions, creating different segmentation maps with different values of $S$.

Figure \ref{fig:supp_cs_ablation} demonstrates segmentation results under three sets of scene regions and style regions. In the first example, the change from $C=8, S=14$ to $C=19, S=25$ results in a more varied stylization result; the walls and ceiling are dissected into smaller scene regions which are given different styles. However, "over-segmentation" of the style image $\mathbf{s}$ will only result in smaller style regions with similar patterns and colors, i.e. it will not create significant changes towards the stylization result, as demonstrated by the results of increasing $C,S$ to $C=41, S=47$. A similar trend can be observed for the second example as well.

\subsection{Injective and surjective mapping}
\label{section:supp_inj_surj}

When computing the mapping $\mathcal{M}$, our current method assumes that $C\leq S$, i.e. $\mathcal{M}$ is injective. Under this assumption, every scene region is matched to a unique style region, which prevents a single local style from dominating the stylization.

However, the computation of our style loss $\mathcal{L}_S$ can take in any arbitrary mapping function. For example, in the case where $C < S$, we can produce a \textit{surjective mapping} where every style region has to be used for stylization.

\begin{figure}[h!]
	\centering
	\includegraphics[width=0.48\textwidth]{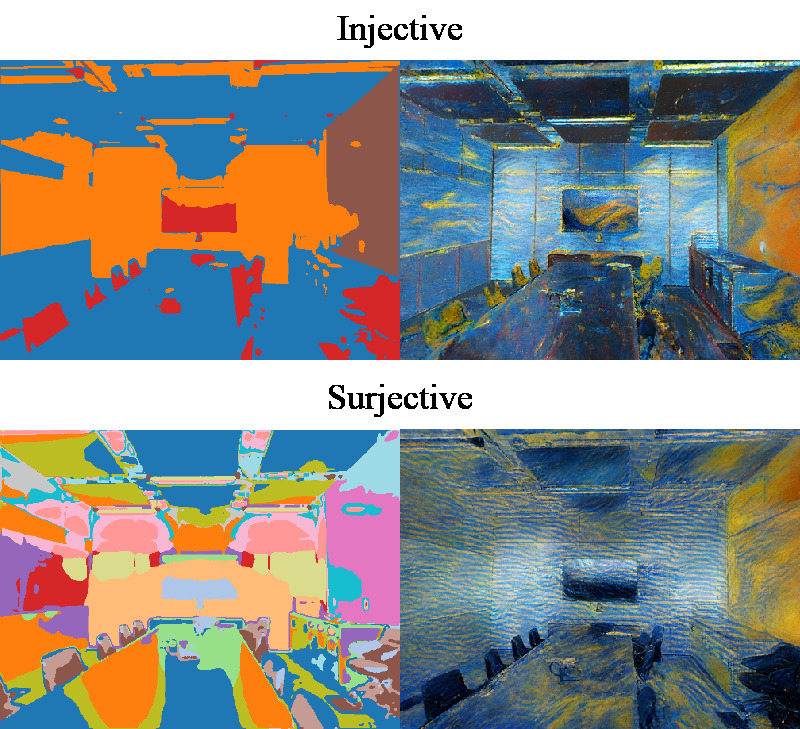}
	\caption{Injective vs. surjective mapping comparison.}
	\label{fig:supp_inj_surj}
\end{figure}

\begin{figure*}[t!]
	\centering
	\includegraphics[width=\textwidth]{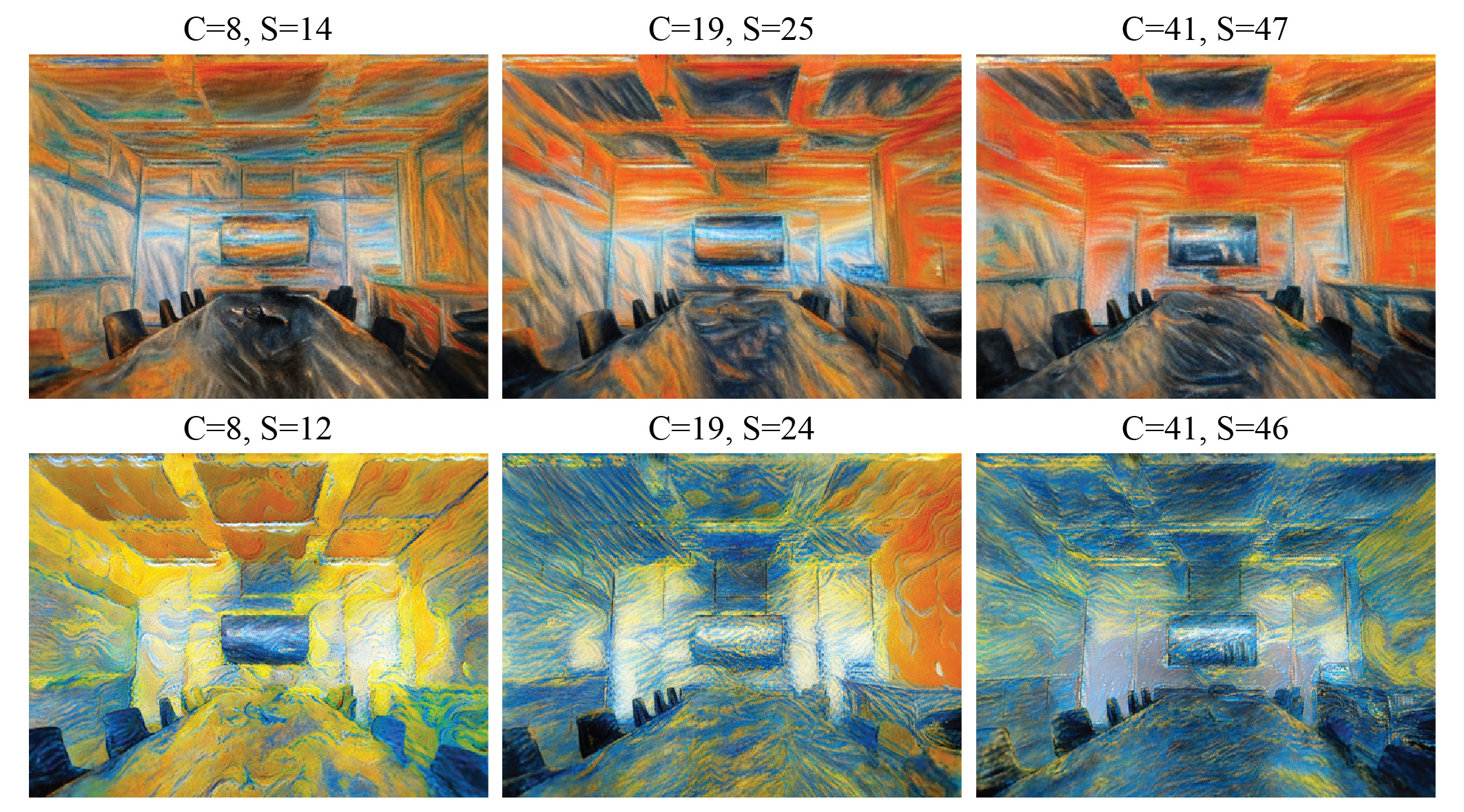}
	\caption{Stylization results by using segmentation maps of different degree of fineness.}
	\label{fig:supp_cs_ablation}
\end{figure*}

To demonstrate this point, we provide an additional ablation experiment comparing between injective mapping and surjective mapping in Fig. \ref{fig:supp_inj_surj}. Under our default injective setting, we have $C=4, S=15$. By reducing the number of training iterations during the segmentation of scene regions, we can increase the number of $C$ from 4 to 26. To generate a surjective mapping with 26 scene regions, we run our current Hungarian algorithm matching to obtain a bijective mapping for 15 scene regions; then run the algorithm again to match the remaining 9 scene regions.

In this example, we observe that the increase in scene regions means a more diverse stylization result; for example, the chairs are no longer constrained to be stylized in the same style as the ceiling. Nevertheless, many scene regions similar in nature are stylized similarly. This arises from the fact that multiple scene regions are now assigned to the same style region, or style regions that have similar patterns and appearances.

One future direction to extended our current method is to improve the automatic matching algorithm such that any general mapping $\mathcal{M}$ can be considered as a candidate.

\begin{figure}[t]
	\centering
	\includegraphics[width=0.4\textwidth]{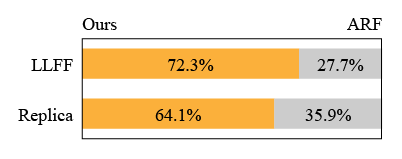}
	\caption{User study results.}
	\label{fig:user_study}
\end{figure}

\begin{figure*}[h!]
	\centering
	\includegraphics[width=\textwidth]{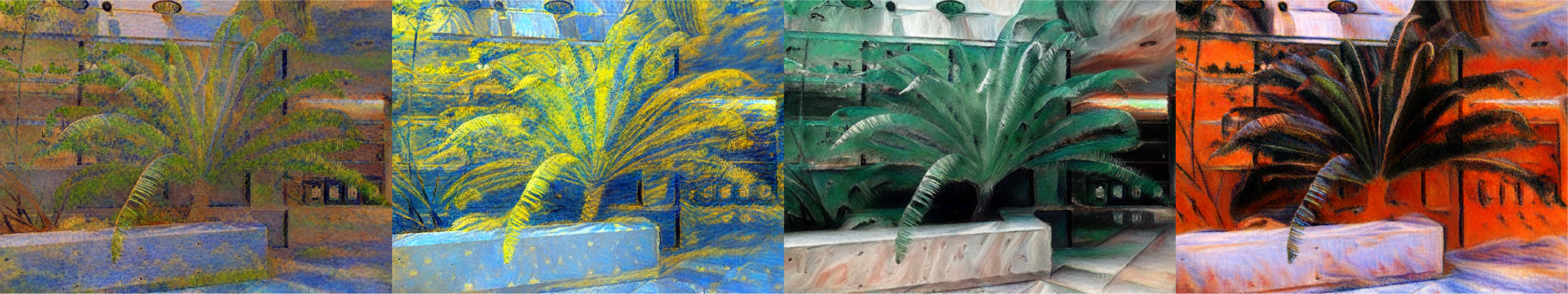}
	\caption{Multiple styles rendered from the same model.}
	\label{fig:supp_multistyle}
	\vspace{0.5in}
\end{figure*}

\begin{figure*}[h!]
	\centering
	\includegraphics[width=\textwidth]{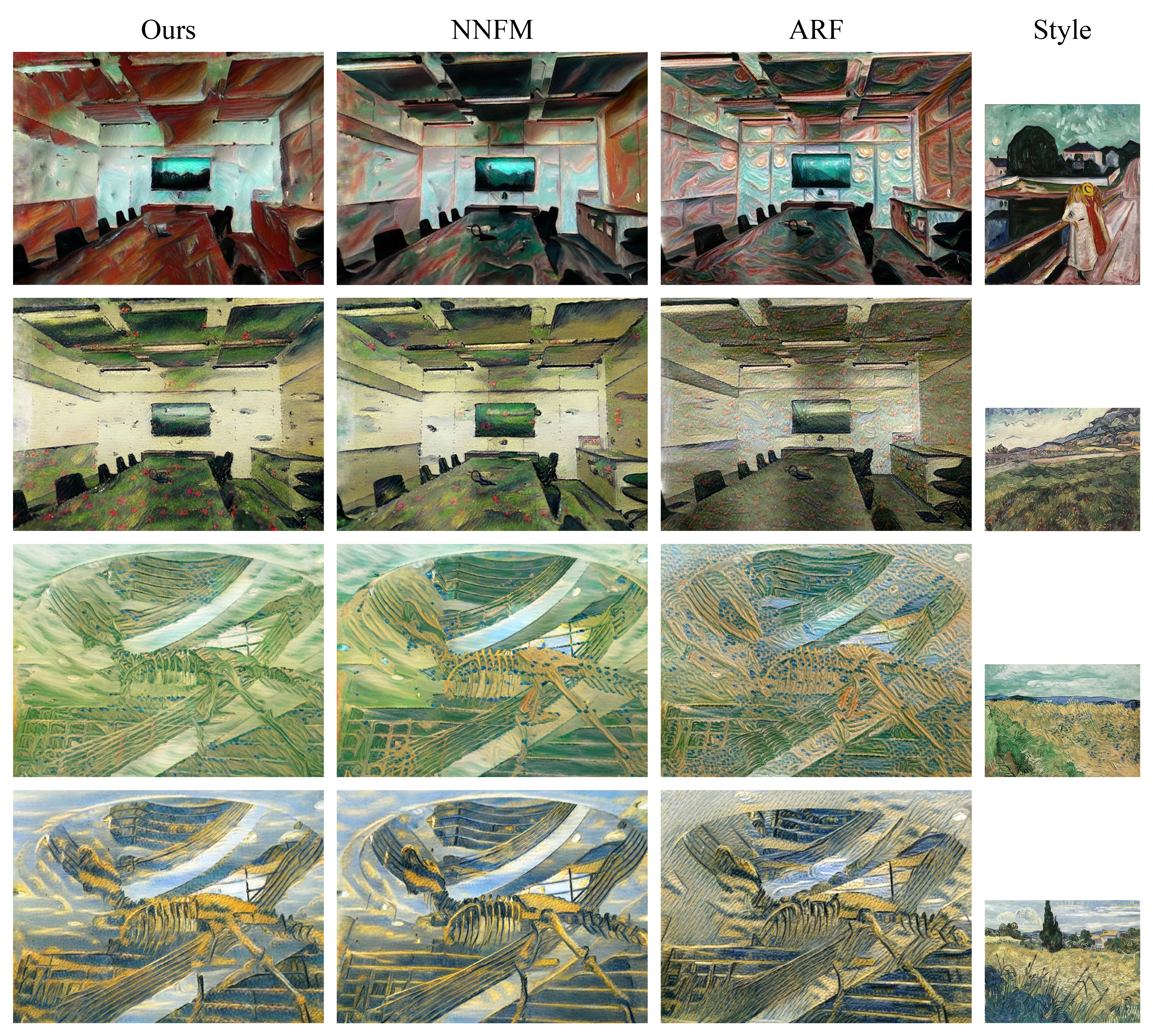}
	\caption{Comparison with and without regional matching. Left column uses regional matching; middle column do not use regional matching, i.e. reduces to the NNFM loss which looks for closest feature vector across entire image; Right column shows result from ARF.}
	\label{fig:supp_nnfm}
\end{figure*}

\section{User study}

We conduct a user study to verify the performance of our proposed method compared with ARF~\cite{zhang2022arf}. The study consists of 20 questions presented in random order; each question consists of two images rendered from a scene stylized by our method; as well as the two ARF-generated images with the same scene and camera pose. The order of the two choices are randomized. In addition, the corresponding ground truth training image and the style images are also shown. The user is asked to select the choice that preserves the content of the ground truth image, and simultaneously appears similar to that of the style image. Out of the 20 questions, 12 of them correspond to images rendered from the \texttt{trex}, \texttt{room} and \texttt{fern} scenes from the LLFF~\cite{mildenhall2019llff} dataset; and 8 of them correspond to images rendered from the \texttt{office3} and \texttt{frl\_apartment3} scenes in the Replica dataset~\cite{replica19arxiv}.

We collected a total of 23 replies and the percentage of picking each method is summarized in Figure \ref{fig:user_study}. The study shows that on average our method is picked at a higher percentage than ARF on both the LLFF and Replica datasets.

\section{Further qualitative results}

We show in this section the results of simultaneously training the stylization of multiple styles within a single hash grid. Figure \ref{fig:supp_multistyle} shows the \texttt{fern} scene from LLFF stylized in four distinct styles. Figure \ref{fig:supp_nnfm} compares the difference between with and without regional matching.

\begin{figure*}[t!]
	\centering
	\includegraphics[width=\textwidth]{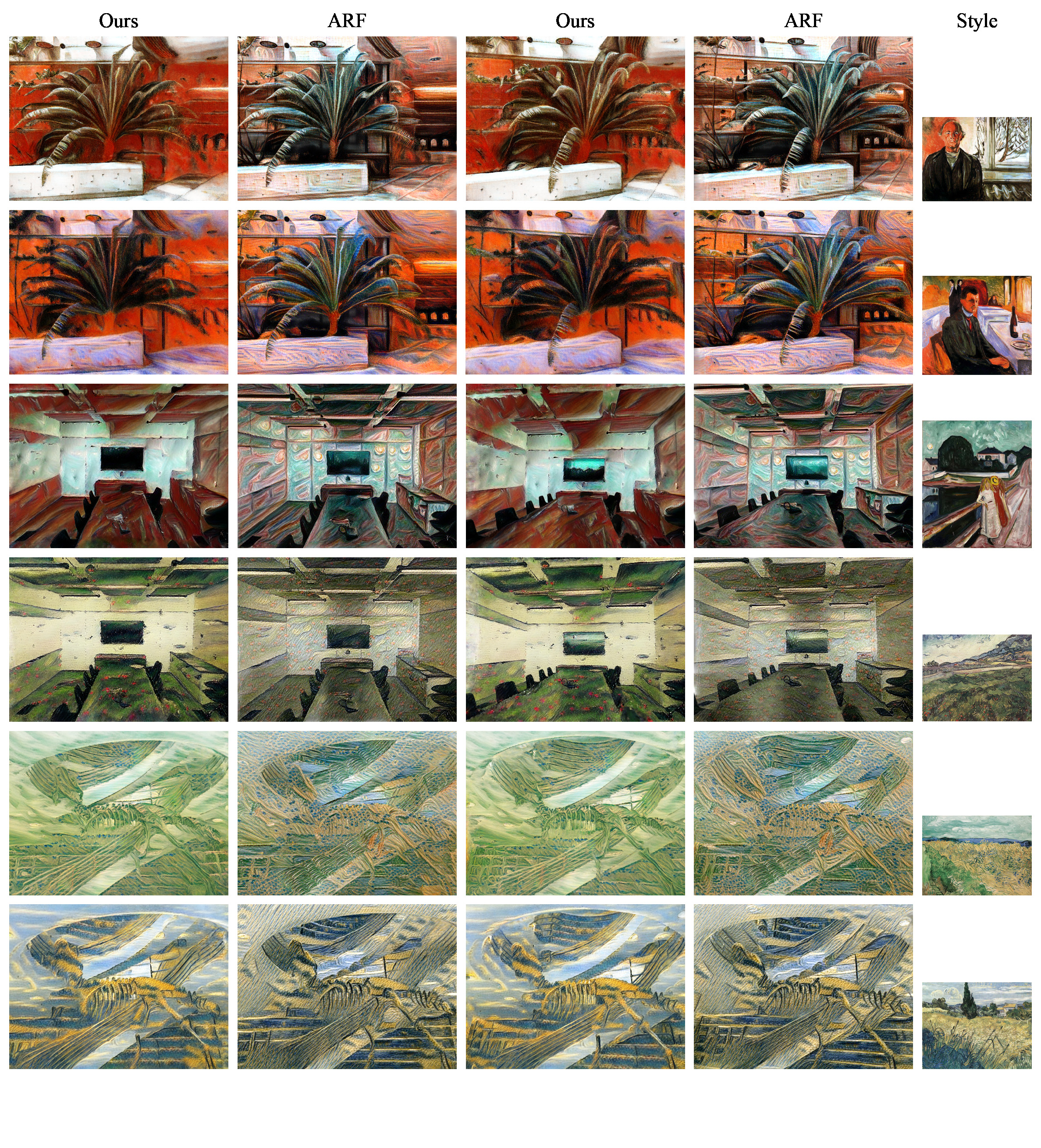}
	\caption{Further qualitative comparison on LLFF dataset.}
	\label{fig:supp_llff}
	\vspace{1.1in}
\end{figure*}

\begin{figure*}[t!]
	\centering
	\includegraphics[width=\textwidth]{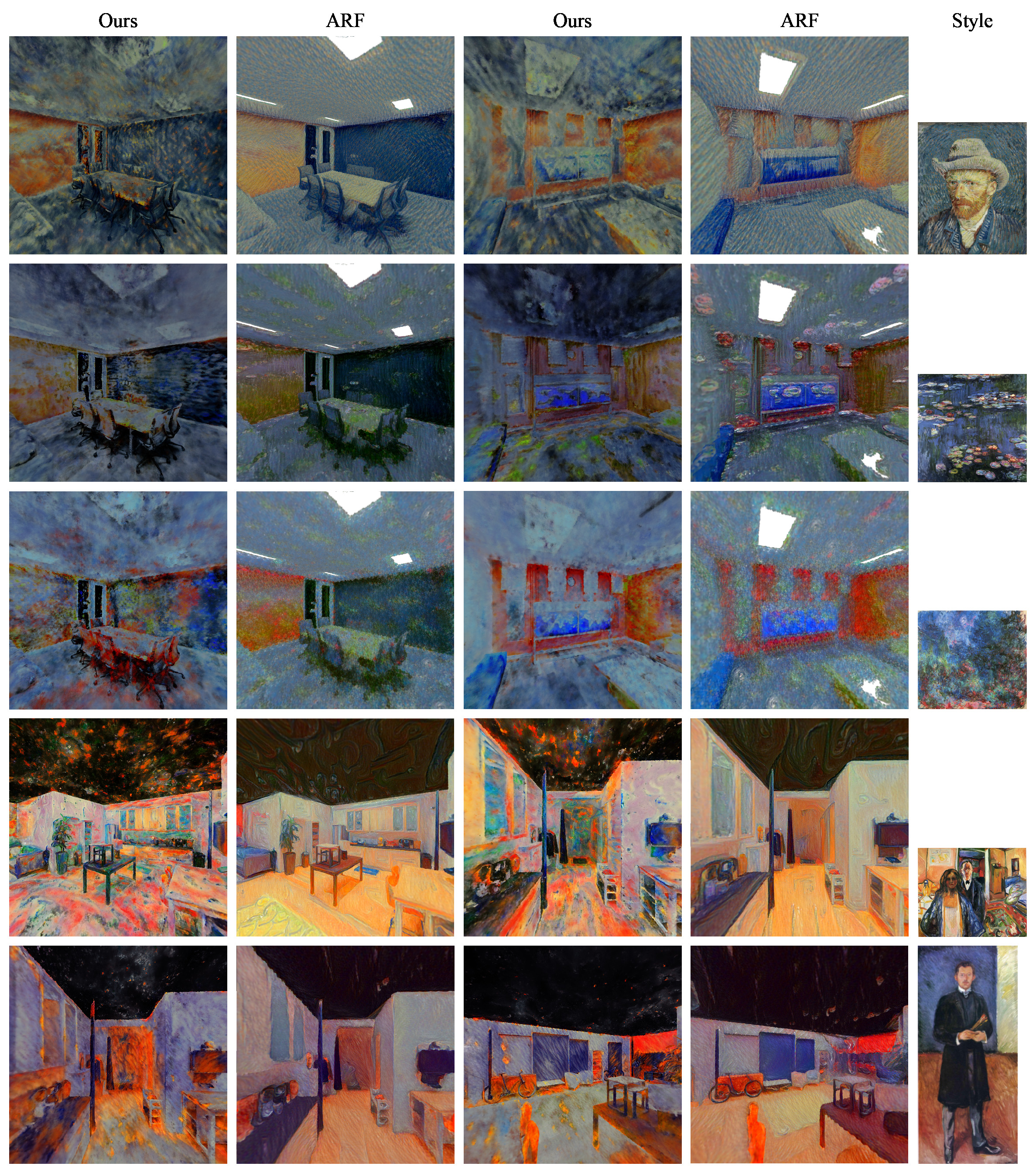}
	\caption{Further qualitative comparison on Replica dataset.}
	\label{fig:supp_replica}
	\vspace{0.8in}
\end{figure*}

We provide further qualitative results for the LLFF dataset in Figure \ref{fig:supp_llff}, and for the Replica dataset in Figure \ref{fig:supp_replica}. Both figures illustrate that our method is less likely to transfer similar, repetitive patterns to the NeRF scene. This is especially the case in low-frequency regions, e.g. concrete walls and bare surfaces.

Our algorithm for matching content-style regions work by assuming that the regions can be paired up in a meaningful sense. However, even in cases where there is little to no correlation between regions from the NeRF scene and style image, our method is able to transfer local patterns on the scene.

\begin{figure*}[t!]
	\centering
	\includegraphics[width=\textwidth]{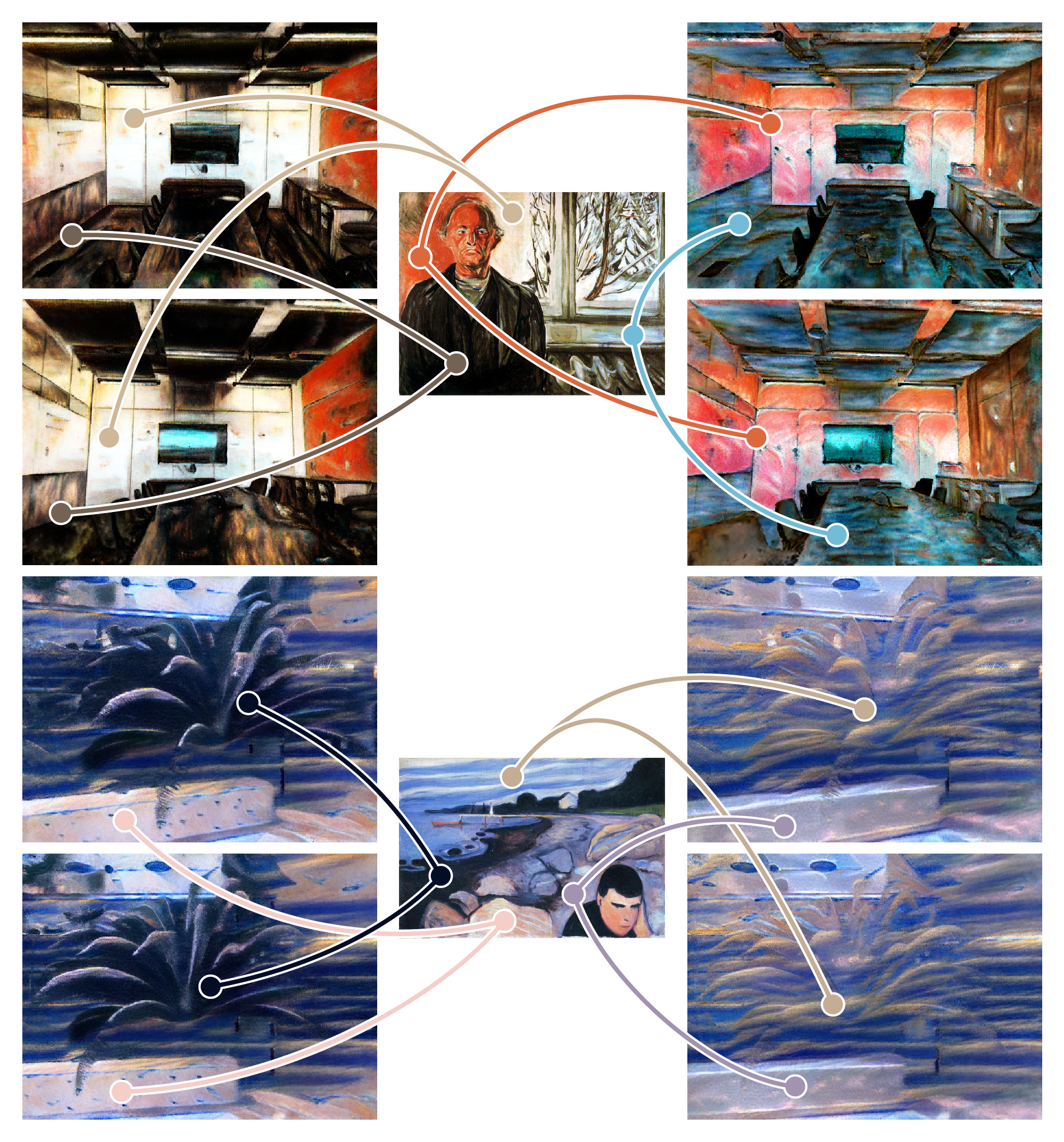}
	\caption{Effect of modifying the pairing between content and style regions. In each example, two content regions have been mapped to two different style regions in the style image (middle column), leading to two completely different stylization results (left and right columns).}
	\label{fig:supp_matching}
	\vspace{0.9in}
\end{figure*}

Finally, we provide two further examples of modifying the pairing between content and style regions in Figure \ref{fig:supp_matching}. We demonstrate that our method can achieve diverse and customizable stylization results via adjusting the pairing.

{\small
\bibliographystyle{ieee_fullname}
\bibliography{egbib}
}

\end{document}